\documentclass[conference]{IEEEtran}
\usepackage{fancyhdr}
\IEEEoverridecommandlockouts

\usepackage{cite}
\usepackage{amsmath,amssymb,amsfonts}
\usepackage{graphicx}
\usepackage{textcomp}
\usepackage{subcaption}
\usepackage{hyperref}
\usepackage{xcolor}
\usepackage{url}
\usepackage{booktabs}
\usepackage{mdframed}

\def\BibTeX{{\rm B\kern-.05em{\sc i\kern-.025em b}\kern-.08em
    T\kern-.1667em\lower.7ex\hbox{E}\kern-.125emX}}

\begin{document}

\title{Scaling Laws for Physics-Aware ACOPF Surrogate Learning}

\author{\IEEEauthorblockN{Yijiang Li\textsuperscript{*}\thanks{\textsuperscript{*}These authors contributed equally to this work.}}
\IEEEauthorblockA{\textit{Argonne National Laboratory}\\
Lemont, USA \\
yijiang.li@anl.gov}
\and
\IEEEauthorblockN{Emon Dey\textsuperscript{*}}
\IEEEauthorblockA{\textit{Argonne National Laboratory}\\
Lemont, USA  \\
edey@anl.gov}
\and
\IEEEauthorblockN{Stefano Fenu}
\IEEEauthorblockA{\textit{Argonne National Laboratory}\\
Lemont, USA \\
sfenu@anl.gov}
\and
\IEEEauthorblockN{Massimiliano Lupo Pasini}
\IEEEauthorblockA{\textit{Oak Ridge National Laboratory}\\
Oak Ridge, USA \\
lupopasinim@ornl.gov}
\and
\IEEEauthorblockN{Teja Kuruganti}
\IEEEauthorblockA{\textit{Oak Ridge National Laboratory}\\
Oak Ridge, USA \\
kurugantipv@ornl.gov}
\and
\IEEEauthorblockN{Kibaek Kim}
\IEEEauthorblockA{\textit{Argonne National Laboratory}\\
Lemont, USA \\
kimk@anl.gov}
}

\maketitle
\thispagestyle{fancy}
\lhead{}
\rhead{}
\chead{}
\lfoot{\footnotesize{
SC26 Workshops, November 15-20, 2026, Chicago, Illinois, USA
\newline 979-8-3195-1221-5/26/\$31.00 \copyright 2026 IEEE}}
\rfoot{}
\cfoot{}
\renewcommand{\headrulewidth}{0pt}
\renewcommand{\footrulewidth}{0pt}

\begin{abstract}
Learning-based surrogates for AC optimal power flow (ACOPF) promise large speedups over classical solvers, but their operational value depends on physical feasibility as much as predictive accuracy. Physics-aware objectives such as the augmented Lagrangian (AL) improve constraint satisfaction at additional per-step cost, yet how this trade-off behaves with scale is uncharacterized. We sweep model and dataset sizes under MSE and AL training, and measure how violation changes with network size. Both objectives improve as power laws: MSE prediction loss depends more on model size than on data, while the AL composite of prediction loss and violation improves comparably along both axes. Violation grows markedly more slowly with network size under AL. On matched hardware and equal training samples, AL reduces violation about $19\times$ with $2.4\times$ the training time and negligible added memory. The training objective shapes not only where a surrogate lands but how its quality evolves with scale.
\end{abstract}


\section{Introduction}
\label{sec:intro}
The electric power grid is among the most critical infrastructure systems in modern society. Its operation is undergoing a profound transformation: the rapid integration of variable renewable generation, distributed energy resources, and electrified transportation and heating loads is increasing both the volume and the volatility of operating conditions that system operators must manage. At the same time, extreme weather events and evolving reliability requirements have raised the stakes of operational decisions, demanding faster and more frequent analysis of a growing space of contingency and planning scenarios. Meeting these demands requires computational tools that can evaluate grid operating decisions at a speed and scale far beyond what traditional workflows were designed to support, making the grid a compelling domain for high-performance scientific machine learning.

AC optimal power flow (ACOPF) is a foundational optimization problem in power system operations, used to determine economically optimal generator 
dispatch subject to nonlinear physical constraints and operational limits~\cite{carpentier-opf-1962, 744492, frank2012optimal_I, frank2012optimal_II}. Modern grid analyses increasingly demand massive what-if studies that require ACOPF to be solved repeatedly at scales where classical nonlinear 
solvers become prohibitive, motivating learning-based surrogates that amortize solver cost through offline training~\cite{pan2022deepopf, fioretto2020predicting, piloto2024canos}. Unlike typical machine learning settings, these surrogates are evaluated on physical feasibility as well as 
predictive accuracy. Predictions with large power balance or thermal limit violations are of limited operational use regardless of how closely they 
match the reference solution~\cite{donti2021dc3, zeng2024qcqp}. Reducing constraint violation is therefore a first-order objective. As these surrogates move toward foundation-model-style deployment across multiple topologies and operating regimes~\cite{hamann2024foundation}, two questions become central for the high-performance computing community responsible for training them: how does surrogate quality scale with model size, dataset size, and compute budget, and how does the choice of training objective, pure regression versus physics-aware losses, change that scaling behavior.


Scaling laws have become a standard tool for characterizing learning systems in language and vision~\cite{kaplan2020scaling, hoffmann2022training}, where 
they predict how model performance improves as a power law in model size, dataset size, and training compute. Their value is predictive: by fitting a 
small number of runs at modest scale, one can forecast the performance a larger model would achieve before committing to training it, and identify 
which resource, capacity or data, is the binding constraint. 

This predictive role is especially consequential for power grid surrogates. Training a foundation-scale ACOPF surrogate requires substantial allocations on leadership-class systems, and without an empirical basis for projecting performance, decisions about model size and data generation rest on extrapolation from small-scale pilots. In scientific machine learning, and in constrained surrogate learning specifically, this picture is far less developed. Physics-aware objectives such as augmented Lagrangian (AL) methods trade additional per-step compute for improved feasibility~\cite{bouchkati2024augmented}, but how this trade interacts with scale remains uncharacterized, leaving the two decisions of how much to scale and what to optimize without an empirical basis to connect them.

In this work, we present a scaling-law study of ACOPF surrogate learning under both physics-aware training (AL) and MSE regression. Building on the LUMINA suite~\cite{jin2026lumina, li2026lumina}, we fit empirical scaling relationships for both predictive error and topology-normalized constraint violation and characterize the scaling behavior of each training regime. 
This informs design decisions for future grid foundation models under varying computational budgets, and for related grid optimization problems that share ACOPF's constraint structure. At the same time, several of our findings can serve as a reference for practitioners deciding when and how to deploy physics-aware training under fixed compute budgets.

Our contributions for this paper are the following, 
\begin{itemize}
    \item \textbf{Scaling laws for constrained ACOPF surrogate learning based on an HGT backbone.} We characterize how each objective's target metric scales with model and dataset size, fitting single-axis and joint scaling relationships, and extend the characterization to how prediction error and constraint violation scale with physical network size across grids spanning three orders of magnitude.
    

    \item \textbf{Benefit and cost of physics-aware training.} Physics-aware training improves feasibility but adds per-step cost. We quantify this 
    trade-off through matched MSE and AL runs on a common platform, reporting training time, memory footprint, and the feasibility gain obtained at equal training samples and at matched prediction loss.
    
    \item \textbf{Implications for grid foundation model design.} We synthesize these results into design considerations for building grid foundation models under fixed computational budgets, considering objective selection and model--data allocation that also serve as a reference for practitioners deploying physics-aware training in resource-constrained settings.
\end{itemize}


\section{Related Work}
\label{sec:relevantworks}

\subsection{Scaling laws for neural networks}
Empirical scaling laws characterize how model performance improves as a power law in model size, dataset size, and training compute. Early work established that generalization error follows predictable power-law trends across domains and architectures~\cite{hestness2017dlscaling}, and \cite{kaplan2020scaling} formalized this for autoregressive language models, showing that loss scales smoothly with parameters, data, and compute over several orders of magnitude. \cite{hoffmann2022training, zhang2024scaling} subsequently revisited the compute-optimal allocation between model and data using iso-FLOP sweeps and a joint fit of the loss as a function of both model and data sizes, concluding that model and data should be scaled in approximately equal proportion. Analogous scaling behavior has been documented in vision~\cite{zhai2022scalingvit} and across generative modeling modalities~\cite{henighan2020scaling}. 

Scaling analyses have more recently been extended to scientific machine learning. \cite{subramanian2023towardsscifm} characterizes scaling and transfer behavior for neural operators on PDE families, finding that pretrained operators transfer across physical systems with favorable data efficiency. Scaling behavior has also been reported for chemical~\cite{frey2023neural} and molecular models. On graph-structured data, which is the natural representation for power networks, scaling behavior is comparatively less well characterized, though recent work has begun to establish neural scaling laws for graph neural networks~\cite{liu2024scalinggraphs, li2025scalinglgraphneural}. 

Two gaps are relevant to our setting. First, existing scientific scaling studies overwhelmingly consider purely supervised objectives; how scaling behavior changes when physical constraints are embedded in the training objective remains largely unexamined. Second, scientific surrogates face a dimension absent in language and vision: the size of the physical system itself. For ACOPF surrogates, network size determines the number of constraints that must be satisfied, yet how constraint satisfaction degrades with network size, and whether the training objective affects that degradation, has not been systematically characterized. This work fits model--data scaling laws under both objectives to address the first gap, and separately characterizes the empirical trend of constraint violation with network size under each objective as a complementary measurement bearing on the second.

\subsection{Learning-Based ACOPF Surrogates}
Learning-based surrogates amortize the cost of repeated ACOPF solves by training on solved instances offline and predicting solutions at inference time. Early approaches use fully connected deep neural networks to learn direct mappings from operating conditions to optimal dispatch or a subset of decision variables~\cite{pan2022deepopf, huang2021deepopfv, pan2023deepopfal}. While these methods achieve substantial speedups over iterative solvers, they tie the learned mapping to a fixed network, limiting reuse across topologies. Hybrid approaches pair a learned model with a classical solver, using network predictions to accelerate the solve while retaining solver accuracy~\cite{dong2020smartpgsim}.

Graph neural networks address the fixed-topology limitation by exploiting the grid's graph structure through message passing, allowing a single model to operate on networks of varying size and connectivity~\cite{owerko2020optimal, gao2023physics, liu2022topology, deihim2024initial, piloto2024canos, yang2024topology}. Power networks are inherently heterogeneous. Homogeneous architectures~\cite{kipf2016semi, velivckovic2017graph, yun2019graph} treat all nodes and edges uniformly and cannot represent this structure explicitly, whereas heterogeneous architectures introduce type-specific projections and relation-specific attention that match the physical semantics of the grid. Among these, the Heterogeneous Graph Transformer (HGT)~\cite{hu2020heterogeneous} combines typed message passing with multi-head attention, using node- and edge-type-dependent projections to distinguish component roles. We adopt HGT as the architecture for this study. Large-scale datasets such as OPFData~\cite{lovett2024opfdata, klamkin2025pglearn}, which provides solved ACOPF instances across multiple representative topologies with perturbed operating points, have made systematic evaluation across networks feasible.

A growing line of work frames these surrogates in foundation-model terms, pretraining a single model across topologies and transferring to new grids~\cite{hamann2024foundation}. Realizing this direction requires understanding how surrogate quality scales with model capacity and training data, which is the question this work addresses.

\subsection{Physics-Aware Learning}
In many scientific and engineering problems, physical laws and operational requirements are expressed mathematically as constraints that a valid solution must satisfy. For ACOPF, these comprise equality constraints encoding the AC power flow physics and inequality constraints encoding generation, voltage, and thermal limits. Refer to Section~\ref{sec:background} for more details on the formulation of ACOPF. Purely supervised regression on solved instances optimizes proximity to reference solutions but provides no mechanism for enforcing these constraints. A broad body of work therefore incorporates constraints directly into the learning process.

The most direct approach augments the training loss with terms penalizing constraint residuals. Physics-informed neural networks embed governing equations as soft penalties~\cite{karniadakis2021physics}, and this idea has been applied to OPF by incorporating power flow residuals and KKT conditions into the objective~\cite{nellikkath2022physics, chen2025physics}. Rather than fixing penalty weights, Lagrangian methods adapt them during training. \cite{fioretto2020predicting} introduces a violation-based Lagrangian (VBL) formulation in which dual variables are updated by ascent steps proportional to observed violations, and augmented Lagrangian (AL) methods~\cite{bouchkati2024augmented} combine multiplier updates with a quadratic penalty term. \cite{kotary2024learning} extends this line into a general framework for learning constrained optimization based on deep augmented Lagrangian methods. These objectives improve feasibility relative to pure regression.

A complementary line of work enforces feasibility through the model architecture or through differentiable procedures rather than through the loss alone. \cite{donti2021dc3} predicts a subset of decision variables and completes the remainder by solving the equality constraints, then applies unrolled gradient-based correction steps to reduce inequality violations. Related approaches include feasibility-oriented architectures for ACOPF~\cite{zeng2024qcqp, hien2025alternative} and post-inference correction and calibration procedures applied to trained surrogates~\cite{wang2024data}. 

Across these families, constraint handling is consistently framed as a question of \emph{how much} feasibility a method achieves at a given training scale. What remains unexamined is how these objectives behave \emph{as scale increases}. We study this question for the AL objective, which prior benchmarking~\cite{jin2026lumina, li2026lumina} identifies as a strong physics-aware objective against MSE regression as the supervised baseline.

\subsection{HPC Perspectives on Training Scientific ML Models}
The computational demands of ACOPF have driven substantial work on hardware-accelerated optimization, including GPU implementations of interior-point and augmented Lagrangian methods~\cite{shin2024accelerating, pacaud2025augmented, montoison2025madncl}, sequential quadratic programming~\cite{li2024gpu}, and multi-period formulations exploiting large-memory GPUs~\cite{shin2024scalable, kim2022accelerated}. Learning-based surrogates represent a complementary use of the same hardware, shifting cost from repeated online solves to a one-time offline training investment. Systems-level techniques such as mixed-precision training~\cite{micikevicius2018mixed} and distributed training strategies become material to feasibility rather than incidental optimizations. Recent work scaling GNNs to billions of parameters for atomistic materials modeling~\cite{li2025scalinglgraphneural} illustrates both the promise and the infrastructure demands of this direction on graph-structured scientific data.

Scaling laws are increasingly used not only as scientific descriptions of learning behavior but as planning instruments: fitted exponents and compute-optimal frontiers allow practitioners to project the return on a given allocation and to choose model and dataset sizes before committing compute. If physics-aware training changes the rate at which feasibility improves with scale, then the objective, as well as being a modeling choice, is a determinant of how compute should be allocated. Characterizing this trade-off for ACOPF surrogates is the aim of this work.

\section{Background}
\label{sec:background}

\subsection{AC Optimal Power Flow Problem}
For a set of buses $\mathcal{N}$, generators $\mathcal{G}$, and transmission lines $\mathcal{L}$, ACOPF aims to minimize the total generation cost function $C(P_g)$:
{\small
\begin{equation}
    C(P_g) = \sum_{g \in \mathcal{G}}(c_{2,g}P_{g}^2 + c_{1,g}P_{g} + c_{0,g}), \label{eq:cost_function}
\end{equation}
}
where $P_{g}$ is the active power output of generator $g \in \mathcal{G}$, and $c_{2,g}$, $c_{1,g}$, and $c_{0,g}$ are cost coefficients, subject to a set of constraints on power balance, generation, voltage, and line flow as follows.

\emph{Power Balance Equations}: The net active and reactive power injection at a given bus $i \in \mathcal{N}$ must equal the power flow out of the bus into the rest of the network:
{\small
\begin{align}
    \sum_{g\in\mathcal{G}_i} P_{g} - P_{d,i} = \sum_{j\in\mathcal{N}_i} P_{ij}(V,\theta), \notag\\
    \sum_{g\in\mathcal{G}_i} Q_{g} - Q_{d,i} = \sum_{j\in\mathcal{N}_i} Q_{ij}(V,\theta), \label{eq:pf1}
\end{align}
}
where $P_{ij}(V,\theta)$ and $Q_{ij}(V,\theta)$ are the standard AC branch-flow functions induced by the bus admittance matrix $Y_{ij} = G_{ij} + kB_{ij}$, where $k^2 = -1$.

\emph{Generation Limits}: Active and reactive power generation must be within the physical limits of each generator:
{\small
\begin{equation}
    P_{g}^{min} \leq P_{g} \leq P_{g}^{\max}, \;\; Q_{g}^{\min} \leq Q_{g} \leq Q_{g}^{\max} \label{eq:genbounds}
\end{equation}
}
\emph{Voltage Limits:} Voltage magnitudes and angles must remain within the operational bounds of each bus:
{\small
\begin{equation}
 V_i^{\min} \leq V_i \leq V_i^{\max}, \;\; \theta_{i}^{\min} \leq \theta_{i} \leq \theta_{i}^{\max} \label{eq:voltage}
\end{equation}
}
\emph{Line Flow Limits}: Power flow on a given transmission line must not exceed the line's thermal ratings:
{\small
\begin{equation}
    P_{ij}^2 + Q_{ij}^2 \leq \left( S_{ij}^{\max} \right)^2. \label{eq:linelimit}
\end{equation}
}

Note that power flow equations \eqref{eq:pf1} can be efficiently solved by traditional methods (e.g., Newton-Raphson); however, ACOPF is an NP-hard optimization problem \cite{bienstock2019strong} with inequality constraints \eqref{eq:genbounds}--\eqref{eq:linelimit}.
In LUMINA, surrogates take operating conditions and topology as input and predict ACOPF decision variables; feasibility is evaluated via the residuals of \eqref{eq:pf1} and \eqref{eq:linelimit}.

\subsection{Training Objectives}
\label{s:losses}

Let $\mathbf{y}$ be the ground-truth ACOPF solution labels and $\hat{\mathbf{y}} = f_\theta(G)$ the model prediction. We define residual functions that measure constraint violations induced by $\hat{\mathbf{y}}$: (i) equality residuals $\mathbf{r}(\hat{\mathbf{y}})$ for \eqref{eq:pf1}, and (ii) inequality residuals $\mathbf{h}(\hat{\mathbf{y}}) \leq \mathbf{0}$ for \eqref{eq:linelimit}.

\paragraph{Pointwise regression (MSE).}
The default objective minimizes squared error (MSE) on solution variables:
{\small
\begin{align}
    L_{\text{MSE}}(\theta) = \mathbb{E}\left[ \|\hat{\mathbf{y}} - \mathbf{y}\|_2^2 \right].
\end{align}
}
This objective measures predictive accuracy but does not explicitly enforce feasibility.

\paragraph{Augmented Lagrangian (AL).}
To encourage feasibility during training, we incorporate constraint residuals using an augmented Lagrangian objective \cite{bouchkati2024augmented} as follows:
{\small
\begin{align}
    L_{\text{AL}}(\theta; \boldsymbol{\lambda}, \boldsymbol{\mu}, \rho)
    &= L_{\text{MSE}}(\theta) 
    + \boldsymbol{\lambda}^T \mathbf{r}(\hat{\mathbf{y}})
    + \frac{\rho}{2}\left\|\mathbf{r}(\hat{\mathbf{y}})\right\|_2^2 \notag \\
    &\quad + \boldsymbol{\mu}^T \max\{\mathbf{h}(\hat{\mathbf{y}}), 0\}
    + \frac{\rho}{2}\left\|\mathbf{h}(\hat{\mathbf{y}})\right\|_2^2,
\end{align}
}
where $\boldsymbol{\lambda}$ and $\boldsymbol{\mu}$ are dual variables associated with equality constraints and inequality constraints, respectively, and $\rho > 0$ is a penalty parameter.
During the training for $\theta$, $(\boldsymbol{\lambda},\boldsymbol{\mu})$ are updated periodically using ascent steps with projection for nonnegativity:
{\small
\begin{equation}
    \boldsymbol{\lambda} \leftarrow \boldsymbol{\lambda} + \rho\, \mathbf{r}(\hat{\mathbf{y}}),\qquad
    \boldsymbol{\mu} \leftarrow \boldsymbol{\mu} + \rho\, \max\left\{\mathbf{h}(\hat{\mathbf{y}}), 0\right\} .
    \label{eq:al_updates}
\end{equation}
}
This objective adaptively reweights constraint satisfaction during training and typically reduces violations under distribution shift.

\section{Experimental Setup}
\label{sec:setup}
We organize our experiments into two studies: a model--data scaling sweep on a single representative topology (Section~\ref{sec:setup:nd}), and a cross-topology sweep characterizing how constraint satisfaction changes with physical network size (Section~\ref{sec:setup:topo}). Both use the data, architectures, and evaluation metrics established in LUMINA~\cite{jin2026lumina, li2026lumina}. LUMINA provides the OPFData pipeline, heterogeneous GNN backbone, and previously tuned AL hyperparameters used in this work, allowing the sweep to isolate the effect of scale without additional tuning. General frameworks for learning constrained optimization, such as NeuroMANCER~\cite{neuromancer2023}, offer alternative implementations of physics-aware objectives; a cross-framework comparison is left to future work.

\subsection{Common Elements}
\paragraph{Data.} We use OPFData~\cite{lovett2024opfdata}, which provides solved ACOPF instances across representative network topologies with 
perturbed load profiles, together with the fixed train/validation/test splits. All supervision comes from solver-optimal ACOPF solutions.

\paragraph{Architecture.} All experiments use the Heterogeneous Graph Transformer (HGT)~\cite{hu2020heterogeneous} backbone. Fixing a single architecture isolates the effect of training objective and scale from architectural confounds. 

\paragraph{Training objectives.} We compare mean squared error (MSE) regression on solution variables against the augmented Lagrangian (AL) 
objective, which augments MSE with constraint residual terms and adaptively updated dual variables and penalty parameter. Both are defined in Section~\ref{sec:background}. AL-specific hyperparameters, including dual step size, penalty parameter, warm-up schedule, and multiplier update frequency, are taken from the hyperparameter optimization previously conducted. This work targets scaling behavior rather than objective tuning, and we therefore perform no additional HPO. Hyperparameters other than model and dataset size are held fixed across all runs.

\paragraph{Metrics.} We report prediction error as MSE between predicted and ground-truth ACOPF solution variables, and physical feasibility as a topology-normalized total constraint violation aggregating power balance and line flow residuals, normalized by $\sqrt{N_{\text{bus}}}$, where $N_{\text{bus}}$ is the number of buses in the topology. Since total violation is an $L_{2}$ norm over residuals whose count scales with $N_{\text{bus}}$, this normalization yields an RMS residual per bus, which is independent of network size by construction.

\subsection{Model--Data Scaling Sweep}
\label{sec:setup:nd}
\paragraph{Topology.} The model--data sweep is conducted on case2000. This choice reflects a trade-off: smaller topologies are learned to near-ceiling accuracy at modest scale and reveal little scaling structure, while the largest systems are significantly more expensive to sweep across a full grid. Case2000 is large enough to exhibit non-trivial scaling behavior in both prediction error and constraint violation while remaining tractable across all configurations per objective.

\paragraph{Scaling axes.} We vary model size $N$ by scaling the hidden dimension, holding the number of layers and attention heads fixed so that capacity varies along a single axis. We sweep six model sizes: 100M, 200M, 300M, 400M, 500M, and 671M parameters and five dataset sizes measured in unique training samples: 48K, 96K, 144K, 192K, and 240K. The sweep spans 6.7× in model size and 5× in dataset size; the fitted exponents describe local trends over this range. Here a \emph{(grid) sample} denotes one solved ACOPF instance: a specific load condition on a given network topology, paired with its solver-optimal solution. This combination of model and dataset sizes yields a grid of 30 configurations per training objective. Models are trained for multiple epochs over each dataset, so samples-seen exceeds the unique sample count; the total samples-seen budget is held fixed across model sizes, so that differences in loss across $N$ reflect capacity rather than differences in training budget.

\paragraph{Training protocol.} Learning rate and remaining optimizer hyperparameters are fixed at previously tuned values, as described in~\cite{jin2026lumina}. We do not use early stopping, so that every configuration consumes its full budget and compute accounting remains directly comparable across the grid. All runs use a single fixed seed. Training is conducted in FP32 throughout. Constraint residuals such as power balance are differences between comparably large injections and flows, so reduced precision risks contaminating the violation metrics under study; FP32 keeps both objectives on equal numerical footing. The effect of mixed-precision training on feasibility is left to future work.

\paragraph{Compute allocation across systems.} MSE runs were executed on Aurora at Argonne Leadership Computing Facility (ALCF) and AL runs on Frontier at Oak Ridge Leadership Computing Facility (OLCF), reflecting machine availability during the study period. Because the two systems differ in GPUs per node, we hold the \emph{global} batch size fixed across all runs, so that the loss values the scaling laws are fit to are directly comparable regardless of the computing systems. Accordingly, our scaling-law fits compare loss values, which are hardware-independent, rather than wall-clock time. To enable a fair comparison of training cost between the two objectives, we additionally train a matched configuration with the MSE objective on Frontier, so that MSE and AL cost measurements come from identical hardware and software environments. All results reported in Section~\ref{sec:results:compute} are drawn from this same-platform comparison.

\subsection{Cross-Topology Sweep}
\label{sec:setup:topo}

To characterize how feasibility changes with physical system size, we train HGT under both MSE and AL on eight topologies spanning three orders of magnitude in network size: case30, case57, case118, case500, case2000, case4661, case6470, and case10000. Each topology is trained under a fixed budget of 15K samples over 40 epochs. Each run uses a 100M parameter model, so that differences across topologies reflect problem difficulty rather than differences in training investment. We then fit empirical trends of prediction loss and topology-normalized constraint violation against network size under each objective. This is a single-axis empirical characterization at fixed model and data budget, not a component of the joint model--data scaling law of Section~\ref{sec:setup:nd}; the two are reported as complementary measurements. 

\subsection{Systems and Implementation}


Aurora nodes provide six Intel Data Center GPU Max 1550 accelerators (12 tiles, 128 GB HBM2e per GPU); runs use 256 nodes. Frontier nodes provide four AMD MI250X accelerators (eight GCDs); runs use 384 nodes. All models were implemented in PyTorch 2.10 with PyTorch Geometric 2.8.

\section{Experimental Results}
\label{sec:results}

\subsection{Model--Data Scaling}
\label{sec:results:nd}

We characterize model--data scaling separately for MSE training and AL training. For MSE, the reported quantity is prediction loss. For AL, we report the composite validation score
{\small
\begin{equation}
L_{\mathrm{AL}}^{\mathrm{comp}} = L_{\mathrm{pred}}+V_{\mathrm{norm}},
\label{eq:al-composite-score}
\end{equation}
}

where \(L_{\mathrm{pred}}\) is the prediction loss and \(V_{\mathrm{norm}}\) is the normalized total constraint violation. This composite score summarizes prediction quality and physical feasibility under the chosen normalization. Individual model and data-scaling relationships are estimated by ordinary least squares in log space, while the joint fits use all available configurations in the model--data grid. Single-axis relationships are shown at representative slices; the joint fits in Equations~\ref{eq:mse-joint} and~\ref{eq:al-composite-joint} use all configurations and are the basis for the reported scaling behavior.

We fit each objective against the quantity it is designed to optimize: prediction loss for MSE, which targets only proximity to the reference solution, and the composite score for AL, which targets prediction accuracy and feasibility jointly. The resulting exponents describe how each objective's own target improves with scale and are not intended as a direct comparison of capacity requirements between objectives.

\begin{figure*}[t]
    \centering
    \begin{subfigure}[t]{0.32\textwidth}
        \centering
        \includegraphics[width=\linewidth]
        {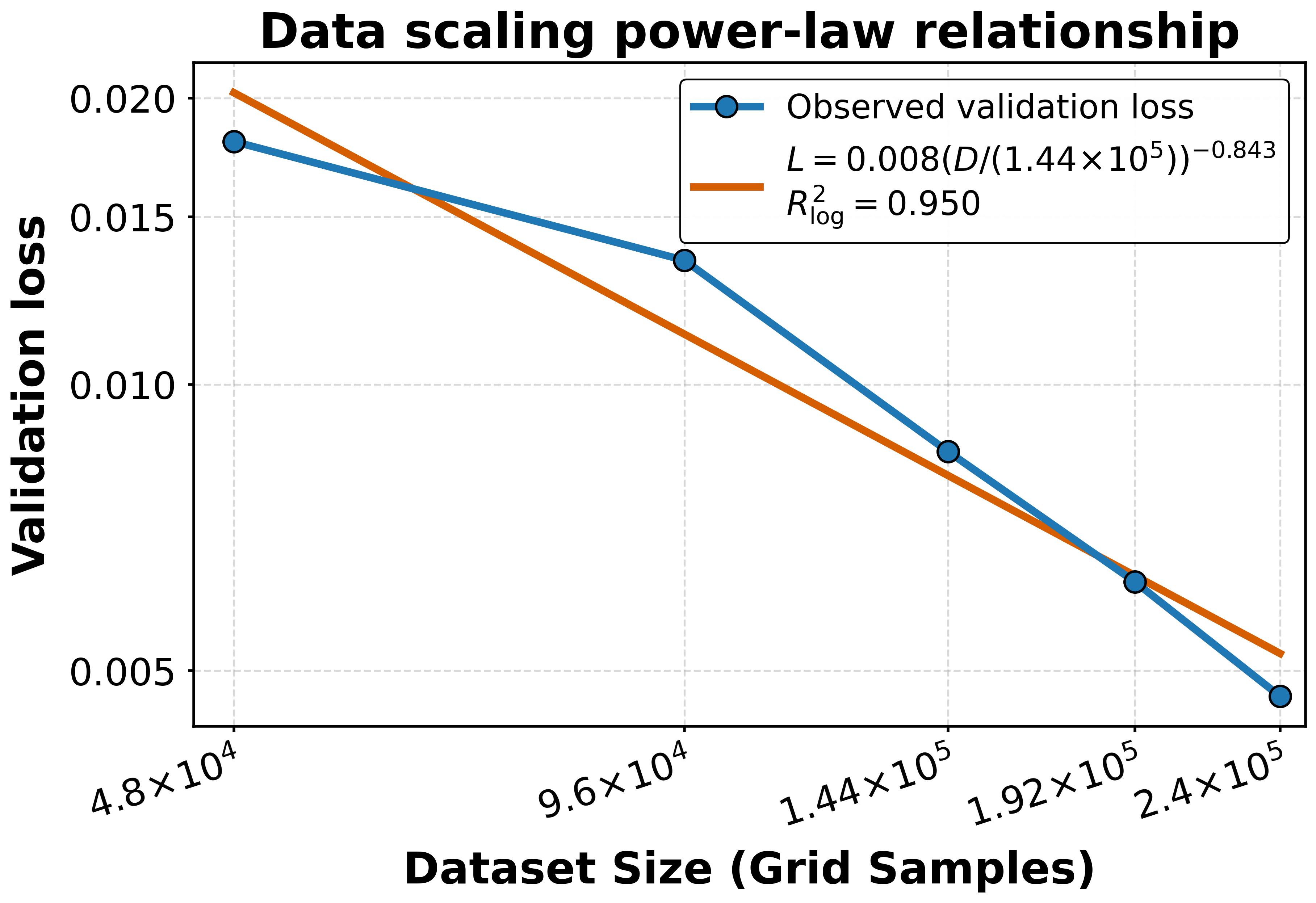}
        \caption{Data scaling at 671M parameters.}
        \label{fig:mse-data-scaling}
    \end{subfigure}
    \hfill
    \begin{subfigure}[t]{0.32\textwidth}
        \centering
        \includegraphics[width=\linewidth]
        {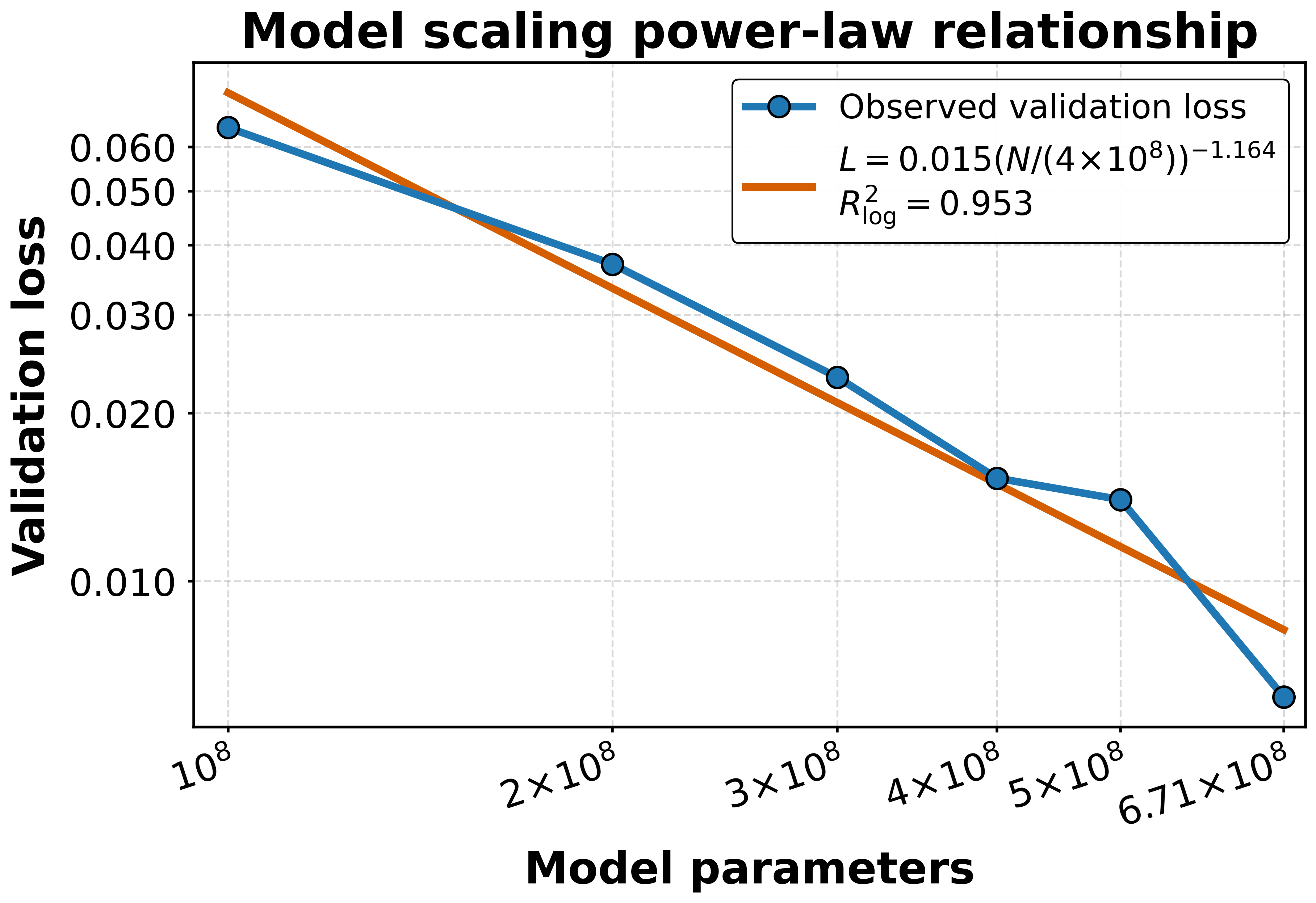}
        \caption{Model scaling at 192K grid samples.}
        \label{fig:mse-model-scaling}
    \end{subfigure}
    \hfill
    \begin{subfigure}[t]{0.32\textwidth}
        \centering
        \includegraphics[width=\linewidth]
        {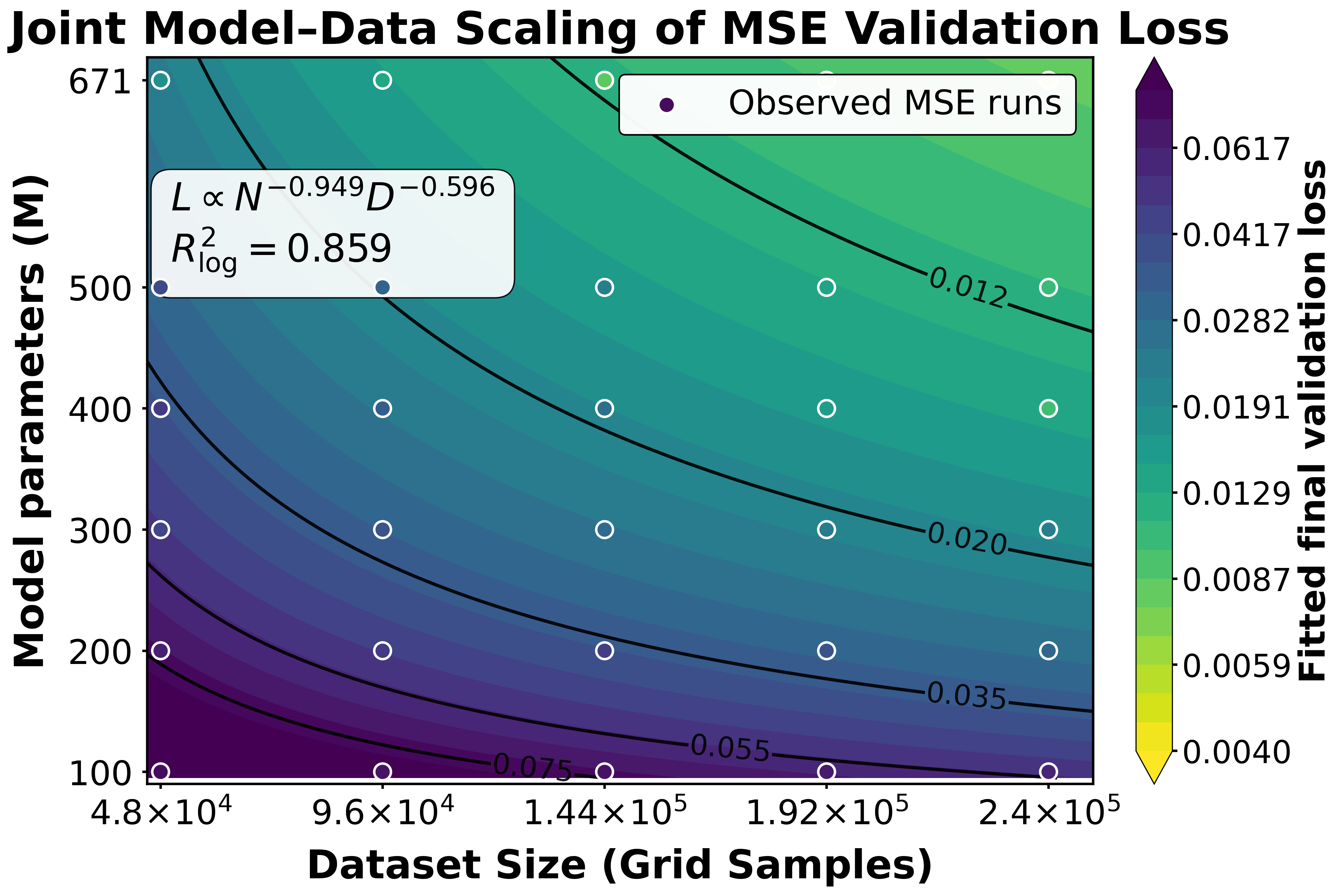}
        \caption{Joint model--data scaling.}
        \label{fig:mse-joint-scaling}
    \end{subfigure}
    \caption{Scaling behavior of MSE validation loss. Points denote measured configurations, lines in (a) and (b) show log--log power-law fits, and the contours in (c) represent the fitted joint loss surface.}
    \label{fig:mse-scaling}
\end{figure*}

\paragraph{MSE scaling.}
Figure~\ref{fig:mse-scaling} characterizes how prediction error responds to additional training examples and model capacity. Let \(N\) denote the number of trainable parameters and \(D\) the number of training samples. With a fixed number of parameters, the data-scaling relationship is
{\small
\begin{equation}
L_{\mathrm{MSE}}(D) = 0.008
\left(
\frac{D}{1.44\times10^{5}}
\right)^{-0.843},
\qquad
R_{\log}^{2}=0.950.
\label{eq:mse-data-scaling}
\end{equation}
}
Increasing the training set from 48K to 240K samples reduces validation loss by approximately \(74\%\). Additional examples reduce finite-data estimation error and provide broader coverage of the operating-condition distribution. For OPF, this exposes the model to a wider range of load, generation, and network-flow states, reducing its dependence on interpolation from a limited set of operating points. The observed power-law behavior is consistent with
the empirical regularity that neural prediction error decreases predictably with dataset size~\cite{kaplan2020scaling,hoffmann2022training}.

With a fixed number of samples, the model-scaling relationship is
{\small
\begin{equation}
L_{\mathrm{MSE}}(N)
=
0.015
\left(
\frac{N}{4\times10^{8}}
\right)^{-1.164},
\qquad
R_{\log}^{2}=0.953.
\label{eq:mse-model-scaling}
\end{equation}
}
Scaling the model from 100M to 671M parameters reduces validation loss by approximately \(90\%\). Greater capacity allows the HGT to represent more detailed interactions among buses, generators, loads, shunts, and transmission elements. The steep fitted exponent indicates that, along this data slice, the remaining prediction error is primarily capacity-limited: the available data can support a richer model than the smaller configurations can express.

To characterize the complete model--data grid, we use a simplified multiplicative scaling form of~\cite{zhang2024scaling}
{\small
\begin{equation}
L(N,D)
=
K N^{-\alpha_N}D^{-\beta_D}.
\label{eq:joint-multiplicative}
\end{equation}
}
The resulting MSE surface is
{\small
\begin{equation}
L_{\mathrm{MSE}}(N,D)
\propto
N^{-0.949}D^{-0.596},
\qquad
R_{\log}^{2}=0.859.
\label{eq:mse-joint}
\end{equation}
}
The fitted surface descends along both axes, showing that model and data scaling provide complementary benefits rather than substituting completely for one another. Its stronger dependence on \(N\) indicates that model capacity is the dominant limiting factor within the measured grid. The nonzero data exponent shows that larger models require sufficient operating-point coverage to realize their predictive capacity.

\begin{figure*}[t]
    \centering
    \begin{subfigure}[t]{0.32\textwidth}
        \centering
        \includegraphics[width=\linewidth]
        {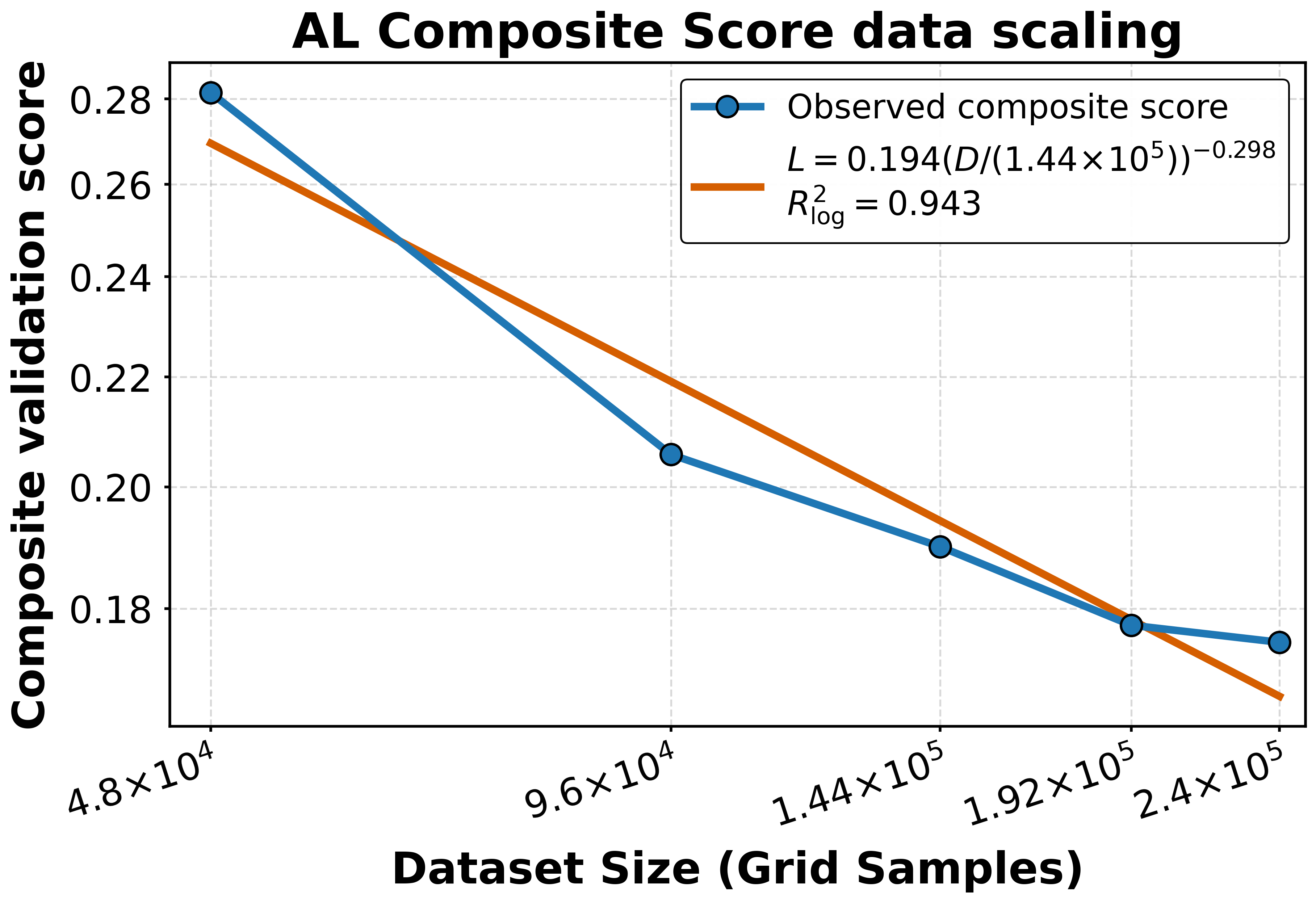}
        \caption{Data scaling at 400M parameters.}
        \label{fig:al-composite-data-scaling}
    \end{subfigure}
    \hfill
    \begin{subfigure}[t]{0.32\textwidth}
        \centering
        \includegraphics[width=\linewidth]
        {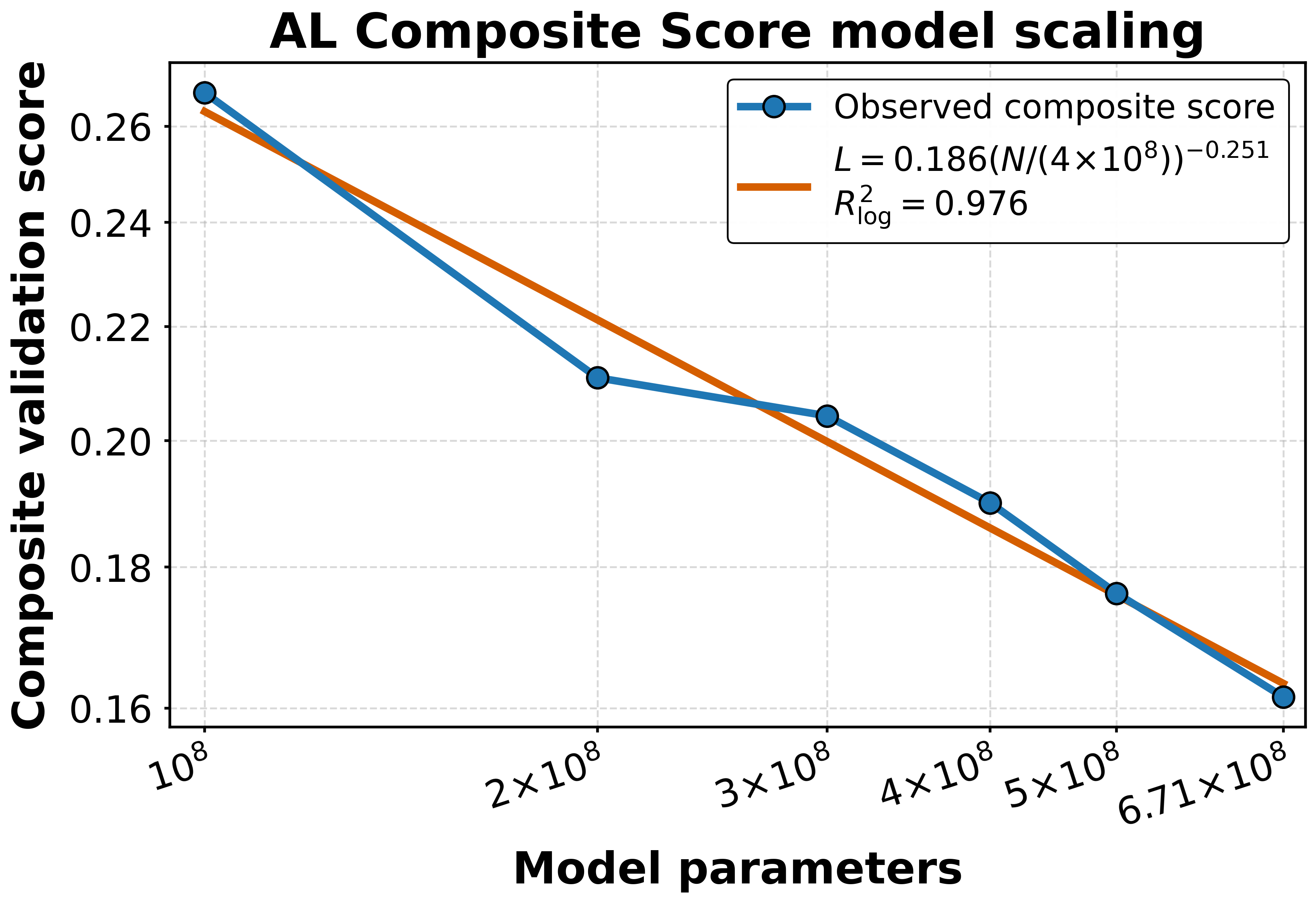}
        \caption{Model scaling at 144K grid samples.}
        \label{fig:al-composite-model-scaling}
    \end{subfigure}
    \hfill
    \begin{subfigure}[t]{0.32\textwidth}
        \centering
        \includegraphics[width=\linewidth]
        {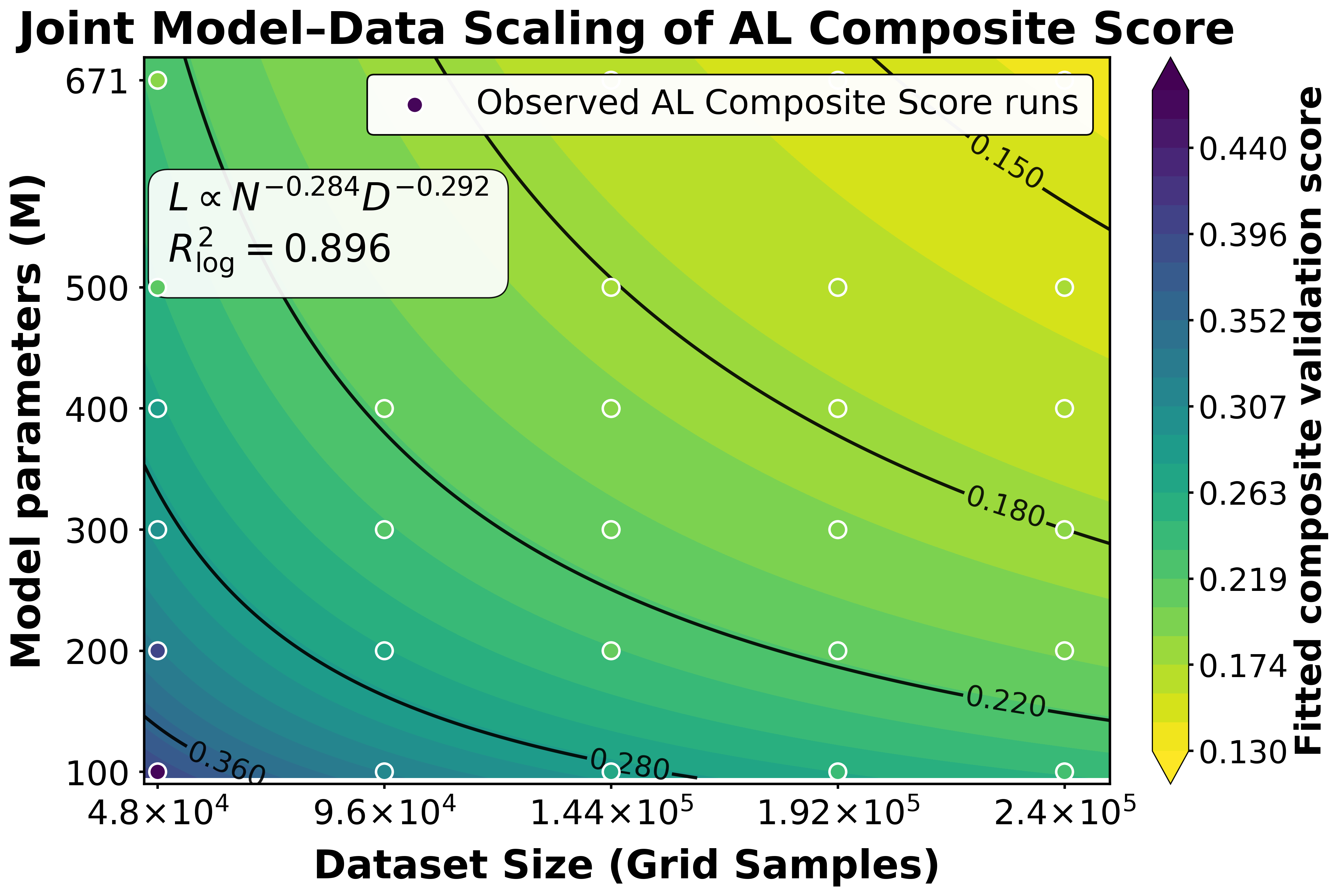}
        \caption{Joint model--data scaling.}
        \label{fig:al-composite-joint-scaling}
    \end{subfigure}
    \caption{Scaling behavior of the AL composite validation score,
    defined as prediction loss plus normalized total constraint violation. Points denote measured configurations, lines in (a) and (b) show log--log power-law fits, and the contours in (c) represent the fitted joint score surface.}
    \label{fig:al-composite-scaling}
\end{figure*}

\paragraph{AL composite-score scaling.}
Figure~\ref{fig:al-composite-scaling} shows how the combined prediction and feasibility measure responds to increasing data and model capacity. Here, the derived data-scaling relationship is
{\small
\begin{equation}
L_{\mathrm{AL}}^{\mathrm{comp}}(D)
=
0.194
\left(
\frac{D}{1.44\times10^{5}}
\right)^{-0.298},
\qquad
R_{\log}^{2}=0.943.
\label{eq:al-composite-data-scaling}
\end{equation}
}

Increasing the dataset from 48K to 240K samples reduces the composite score by approximately \(38\%\). Additional samples serve two related purposes: they improve estimation of the input--output mapping and expose the model to a wider variety of constraint-active operating conditions. The latter is particularly important for physical feasibility because line-limit and balance residuals may be most informative near restricted or comparatively rare regions of the operating distribution. The observed model-scaling relationship is
{\small
\begin{equation}
L_{\mathrm{AL}}^{\mathrm{comp}}(N)
=
0.186
\left(
\frac{N}{4\times10^{8}}
\right)^{-0.251},
\qquad
R_{\log}^{2}=0.976.
\label{eq:al-composite-model-scaling}
\end{equation}
}
Increasing model capacity from 100M to 671M parameters reduces the composite score by approximately \(40\%\). Under the composite objective, additional capacity must support both accurate OPF prediction and the structure needed to reduce physical violations. The strong log-space fit indicates that this combined task benefits systematically from model growth rather than improving only at a particular parameter scale.

Applying Equation~\ref{eq:joint-multiplicative} to all available AL configurations gives
{\small
\begin{equation}
L_{\mathrm{AL}}^{\mathrm{comp}}(N,D)
\propto
N^{-0.284}D^{-0.292},
\qquad
R_{\log}^{2}=0.896.
\label{eq:al-composite-joint}
\end{equation}
}
The nearly equal model and data exponents indicate that neither capacity nor operating-point coverage is a clearly dominant bottleneck for the composite score within the evaluated regime. More parameters improve the model's ability to represent predictive and feasibility-related structure, while more samples provide the diversity required to learn that structure across operating conditions.



We assessed the stability of the joint fits with 2,000 bootstrap resamples of the 30 grid configurations, refitting Equation~\ref{eq:joint-multiplicative} to each resample and reporting percentile 95\% confidence intervals (Table~\ref{tab:bootstrap-exponents}). These intervals capture fit uncertainty across configurations, not seed-to-seed variability. Under MSE, the bootstrap interval for $\alpha_N - \beta_D$ excludes zero, supporting model-dominant scaling; under AL it includes zero, consistent with balanced dependence on model and data. 
\begin{table}[h]
\centering
\caption{Bootstrap estimates of the joint scaling exponents.}
\label{tab:bootstrap-exponents}
\resizebox{\columnwidth}{!}{%
\begin{tabular}{llccc}
\hline
Training & Fitted quantity
& $\alpha_N$ [95\% CI]
& $\beta_D$ [95\% CI]
& $R^2_{\log}$ \\
\hline
MSE & Prediction loss
& 0.949 [0.748, 1.157]
& 0.596 [0.397, 0.811]
& 0.859 \\
AL & Composite score
& 0.284 [0.225, 0.350]
& 0.292 [0.201, 0.374]
& 0.896 \\
\hline
\end{tabular}%
}
\end{table}

As a sensitivity analysis, we also fitted the additive Chinchilla-style~\cite{hoffmann2022training} form \(L(N,D)=E+A N^{-\alpha_N}+B D^{-\beta_D}\). Across the measured model-data grid, the fitted value of $E$ was typically at or near zero, and the exponent estimates had broad bootstrap intervals. This indicates that a stable irreducible floor was not identifiable from the present grid because loss continues to decrease across the measured range. Therefore, the exponents in Equations~\ref{eq:mse-joint} and~\ref{eq:al-composite-joint} should be read as effective local exponents over the tested range.

\subsection{Cross-Topology Scaling}
\label{sec:results:topo}

\begin{figure}[t]
    \centering
    \includegraphics[width=0.85\linewidth]{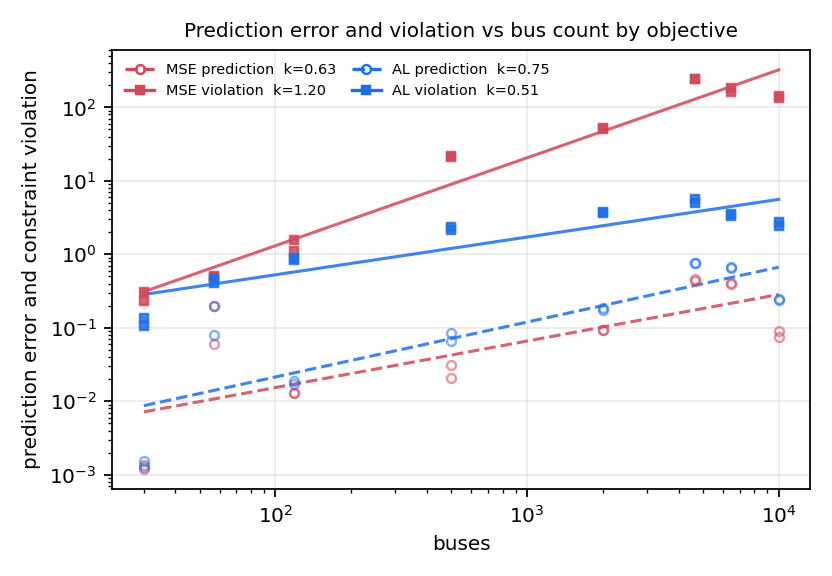}
    \caption{Constraint violation and prediction error as a function of network size (bus count), from case30 to case10000, across AL and MSE objectives with a 100M parameter model.}
    \label{fig:violation_vs_topo}
\end{figure}

Figure~\ref{fig:violation_vs_topo} shows topology-normalized constraint violation and prediction error as a function of network size across case30, case57, case118, case500, case2000, case4661, case6470, and case10000, under both MSE and AL at fixed model and data budget. All reported values for cross-topology experiments are means across two repeated seeds. Fitting $\log(\text{violation})$ against $\log(\text{buses})$ gives $k = 1.196 \pm 0.064$ under MSE and $k = 0.513 \pm 0.068$ under AL, so the violation exponent is lower by $\Delta  k ~\approx 0.68$ under AL. Physics-aware training confers a benefit that grows with physical system size, which is critical for foundation models intended to span heterogeneous networks.

\begin{figure}[t]
    \centering
    \includegraphics[width=0.85\linewidth]{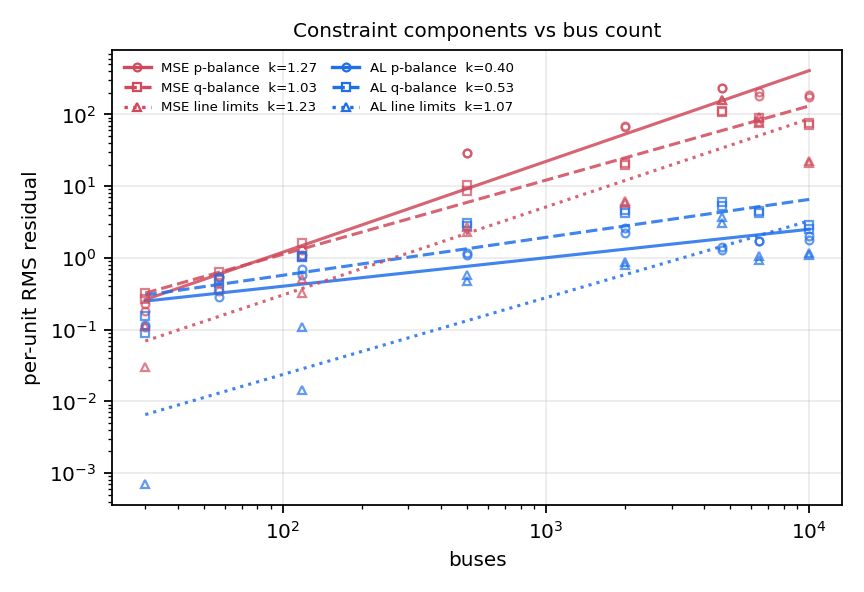}
    \caption{Constraint violation broken down by constraint type, as a function of network size, from case30 to case10000, across AL and MSE objectives with a 100M parameter model.}
    \label{fig:constraint_type_vs_topo}
\end{figure}

Decomposing by constraint type (Figure~\ref{fig:constraint_type_vs_topo}) shows that the gap is largest on active power balance ($k$: $1.266 \to 0.397$, $\Delta k = 0.87$), roughly half as large on reactive power balance ($1.034 \to 0.529$), and almost absent on line limits ($1.228 \to 1.071$). This last is largely because line limit violations were not found to be strongly correlated with network size in general, and depend more on local features of network topology that vary independently of size (constraint specifics, susceptance, etc.). These results show that a physics-aware loss slightly trades off prediction error to achieve better overall constraint violation performance. This suggests a curriculum in which MSE training establishes prediction accuracy before a physics-aware loss is introduced to improve feasibility metrics.  Absolute violation on the largest grids remains non-negligible even under AL, so operational use would pair such surrogates with a feasibility-restoration step or use them as solver warm starts, complementary to hybrid approaches such as~\cite{dong2020smartpgsim}.

\subsection{Benefit and Cost of Physics-Aware Training}
\label{sec:results:compute}

Table~\ref{tab:mse-al-combined} compares MSE and AL training on the matched same-platform runs at 671M parameters and 192K samples, so that cost 
measurements for the two objectives are directly comparable. 
\begin{table}[h]
\centering
\caption{Comparison of MSE and AL training under equal-sample and approximately matched-prediction-loss settings, for the
671M model on the 192K training shard.}
\label{tab:mse-al-combined}
\resizebox{\columnwidth}{!}{%
\scriptsize
\begin{tabular}{lrrrr}
\toprule
& \multicolumn{2}{c}{Equal training samples}
& \multicolumn{2}{c}{Matched prediction loss} \\
\cmidrule(lr){2-3}\cmidrule(lr){4-5}
Metric & MSE & AL & MSE & AL \\
\midrule
Prediction loss
    & 0.1104 & 0.1456
    & 0.1568 & 0.1582 \\
Training samples processed
    & 5.806M & 5.806M
    & 0.968M & 3.484M \\
Elapsed time (h)
    & 0.993 & 2.394
    & 0.192 & 1.832 \\
Normalized total violation
    & 0.3794 & 0.0198
    & 0.4851 & 0.0165 \\
Total violation
    & 66.290 & 3.452
    & 84.758 & 2.876 \\
P-balance violation
    & 57.311 & 2.347
    & 56.459 & 1.435 \\
Q-balance violation
    & 73.582 & 4.106
    & 104.569 & 3.676 \\
Line-limit violation
    & 6.266 & 0.829
    & 9.851 & 0.687 \\
Peak allocated memory (GiB)
    & 25.995 & 26.024
    & 25.995 & 26.024 \\
\bottomrule
\end{tabular}%
}
\end{table}

We report two settings: an equal-sample comparison, in which both objectives process the same number of training samples, and a matched-prediction-loss comparison, in which each objective is trained until it reaches approximately the same prediction loss. The first isolates the per-step cost of physics-aware training; the second reveals what AL costs to reach parity on the metric MSE directly optimizes.


At the same 5.806 million processed samples, AL was approximately \(2.41\times\) slower than MSE. Its prediction loss was 31.9\% higher, but its normalized total violation was \(19.2\times\) lower, representing a 94.8\% reduction, with P-balance, Q-balance, and line-limit violations reduced by \(24.4\times\), \(17.9\times\), and \(7.6\times\), respectively. Peak allocated memory increased by only about 0.029 GiB (30 MiB, or 0.11\%), indicating no practically significant memory overhead from AL. 

At an approximately matched prediction loss of 0.157, AL reduced normalized total violation by 96.6\% (\(29.5\times\)) with P-balance, Q-balance, and line-limit violations reduced by \(39.3\times\), \(28.4\times\), and \(14.3\times\), respectively. Note that the matched-loss MSE model is taken early in training, so its violation (0.4851) is higher than that of the fully trained MSE model (0.3794); within the training budget, AL does not reach MSE's best prediction loss of 0.1104. Reaching this prediction-loss region required AL to process \(3.6\times\) more samples and required \(9.55\times\) more elapsed time. Peak allocated memory remained within 30 MiB of MSE, so the computational tradeoff of physics-aware training is primarily in training time rather than accelerator memory capacity.


Holding prediction accuracy fixed, AL thus reduces violation by nearly 30× relative to an MSE model at the same prediction loss, for roughly an order of magnitude more training time: a trade that favors AL wherever training is a one-time cost and the model is deployed repeatedly.

\section{Implications for Grid Foundation Model Design}
\label{sec:discussions}

Our results bear on several decisions in planning a large-scale ACOPF surrogate.

\paragraph{Neither capacity nor data is saturated within the tested range.} Across the evaluated regime, increasing either model or dataset size continues to improve performance under both objectives, with each objective's target metric improving at its own rate. Under MSE, the joint relationship is more sensitive to model size than to data, whereas under AL both additional capacity and additional data continue to improve the combined objective. For practitioners this means that at these scales, additional gains should remain available by scaling compute along either axis and the most effective allocation depends on the target metric.

\paragraph{The training objective is a scaling decision as well as a modeling one.} 
Which resource gives the better return depends on the metric being targeted. Because the objective determines that metric, objective choice bears directly on compute budget planning. It also raises a question about how surrogate model performance is typically reported, as such models are typically compared under a fixed training budget or a fixed operating point. This mode of comparison can obscure the way model behavior changes with scale, and this must be taken into account when training physics-aware models.

\paragraph{Feasibility and accuracy need not scale together.} In the cross-topology study, where prediction error and constraint violation are tracked separately, we can ask whether scale buys accuracy and feasibility at the same rate. Our results suggest that the answer depends on objective choice: in general, predictive accuracy improvements from scaling do not imply commensurate improvements in constraint violation.


\paragraph{Physical system size is a distinct scaling axis.} At a fixed model and data budget, prediction error grows with network size comparably under both objectives, while the violation exponent is about 0.68 lower under AL than under MSE from case30 to case10000 (Figure~\ref{fig:violation_vs_topo}). Under MSE, a budget adequate for medium networks therefore does not translate to large ones, and training budgets must scale with network size regardless of accuracy targets. Physics-aware training is most valuable for models that have a violation target and  either directly target large networks or cover a broad range of network sizes. The benefit is non-uniform across constraint categories (Figure~\ref{fig:constraint_type_vs_topo}) and should be characterized per constraint family.


\paragraph{Compute considerations.} The cost of physics-aware training is borne almost entirely in training time: at equal training samples, AL required $2.4\times$ the training time of MSE, while peak memory differed by about 0.1\% (Section~\ref{sec:results:compute}). Since memory typically caps trainable model size, AL leaves that ceiling unchanged. Although AL adds to the training cost that motivates scaling-law planning, training is a one-time offline cost amortized over repeated inference, the axis scaling programs can most readily absorb. Where budget is limiting, much of AL's feasibility benefit may be obtainable by training under MSE to an accuracy target and introducing the physics-aware objective thereafter.

\section{Conclusion}
\label{sec:conclusion}


This work presented a scaling-law study of ACOPF surrogate learning, comparing AL training against MSE regression and characterizing how constraint violation grows with network size. Across 100M–671M parameters and 48K–240K samples on case2000, both objectives improve as power laws in model and dataset size. The jointly fitted exponents $(\alpha_N, \beta_D)$ are $(0.949, 0.596)$ for MSE prediction loss and $(0.284, 0.292)$ for the AL composite score (95\% bootstrap CIs in Table~\ref{tab:bootstrap-exponents}): MSE prediction loss depends more strongly on model capacity, while the AL composite improves at comparable rates with capacity and data. At equal training samples, AL reduced normalized violation by about $19\times$ at $2.4\times$ the training time; at approximately matched prediction loss, against an MSE checkpoint taken early in training, the reduction was about $30\times$. Across topologies, violation grows with network size at an exponent of 1.196 under MSE but only 0.513 under AL, while prediction error grows comparably under both. Accuracy and feasibility therefore need not scale at the same rate, and the choice of objective affects scaling behavior, not just final performance. These findings inform design decisions for future grid foundation models under varying computational budgets.

Several directions would extend this study. The model–data sweep covers a single architecture family, topology and random seed, over $6.7\times$ in model size and $5\times$ in dataset size, so the fitted exponents describe local trends over the tested range; our bootstrap intervals capture fit uncertainty rather than seed-to-seed variability. Extending the comparison to other GNN architectures and to other constraint-handling approaches, including VBL, hard-constraint architectures, or post-hoc correction, would show how broadly these scaling patterns apply, as would validation on operational rather than synthetic grid data and the effect of mixed-precision training on feasibility. A further direction is to move beyond a single task. A grid foundation model would serve a family of related optimization problems, and whether scaling behavior established for ACOPF transfers to multi-task training or whether task diversity itself becomes a scaling axis alongside model and data remains open.

\section*{Acknowledgment}
This work was supported by the U.S. Department of Energy, Office of Science, Advanced Scientific Computing Research, under Contract DE-AC02-06CH11357. This material is based upon work supported by the U.S. Department of Energy, Office of Critical Minerals and Energy Innovation (CMEI), under Contract No. DE-AC02-06CH11357. This material is based upon work supported by Laboratory Directed Research and Development (LDRD) funding from Argonne National Laboratory, provided by the Director, Office of Science, of the U.S. Department of Energy under contract DE-AC02-06CH11357. An award of computer time was provided by the ASCR Leadership Computing Challenge (ALCC) program. This research used resources of the Argonne Leadership Computing Facility, which is a U.S. Department of Energy Office of Science User Facility operated under contract DE-AC02-06CH11357. This research used resources of the Oak Ridge Leadership Computing Facility at the Oak Ridge National Laboratory, which is supported by the Office of Science of the U.S. Department of Energy under Contract No. DE-AC05-00OR22725. 


\bibliographystyle{IEEEtran}
\bibliography{reference}

\end{document}